\documentclass[runningheads]{llncs}
\usepackage[T1]{fontenc}
\usepackage{graphicx}
\usepackage{booktabs}
\usepackage[misc]{ifsym}

\usepackage{mwe}
\usepackage{cite}
\usepackage{amsmath} %mathematical package
\usepackage{tablefootnote}
\newcommand{\norm}[1]{\left\lVert#1\right\rVert}
\begin{document}

\title{Compressing CNN models for resource-constrained systems by channel and layer pruning}

\titlerunning{Compressing CNN models for resource-constrained systems}
% If the full title of your paper is short enough to also fit in the running head, you can omit the abbreviated paper title here. You can check as follows: if you comment out the \titlerunning line, something will appear in the header of all odd-numbered pages of your PDF from page 3 onward. This something is either the full title (in which case all is well), or the error message "Title Suppressed Due to Excessive Length". If this error message appears, you're going to want to provide an abbreviated title within the \titlerunning command, because if you won't do it, Springer will do it for you.

%N.B.: Author information (both in the \author{} and \authorrunning{} command) should only be present in the Camera-Ready Version of your paper. The version that you initially submit for review, ought to be double-blind. So, when initially submitting your paper, use:
% \author{Author information scrubbed for double-blind reviewing}
\author{Ahmed Sadaqa\inst{1}, Di Liu \inst{2}
% You may leave out the orcidID information, if you want to.
% Use \corr to indicate the corresponding author. Note the spacing around the \corr command. Only one author can be the corresponding author.
}
%N.B.: comment out the \authorrunning{} command for the double-blind version of your paper submitted for review. Later, if your paper is accepted, use the command for the Camera-Ready Version.
\authorrunning{A. Sadaqa and D. Liu}
% First names are abbreviated in the running head.
% If there is one author, write 'A.L. Benjamin'.
% If there are two authors, write 'A.L. Benjamin and C.C. Broadus Jr.'
% If there are more than two authors, '[...] et al.' is used.

% \institute{Fictional Southern University, Savannah GA 31404, USA \email{\{a.l.benjamin,a.a.patton\}@fsu.fake}
% \and
% Fictional West Coast University, Long Beach CA 90840, USA \email{ccb@fwcu.fake}
% \and
% Secondary European Affiliation, Tiergartenstr. 17, 69121 Heidelberg, Germany
% \email{lncs@springer.com}}

\institute{Norwegian University of Science and Technology, NO-7491 Trondheim, Norway 
\email{ahmedsad@stud.ntnu.no}
\and
Norwegian University of Science and Technology, NO-7491 Trondheim, Norway  \email{di.liu@ntnu.no}
}
\maketitle              % typeset the header of the contribution

\begin{abstract}
Convolutional Neural Networks (CNNs) have achieved significant breakthroughs in various fields. However, these advancements have led to a substantial increase in the complexity and size of these networks. This poses a challenge when deploying large and complex networks on edge devices. Consequently, model compression has emerged as a research field aimed at reducing the size and complexity of CNNs. One prominent technique in model compression is model pruning. This paper will present a new technique of pruning that combines both channel and layer pruning in what is called a "hybrid pruning framework". Inspired by EfficientNet, a renowned CNN architecture known for scaling up networks from both channel and layer perspectives, this hybrid approach applies the same principles but in reverse, where it scales down the network through pruning. Experiments on the hybrid approach demonstrated a notable decrease in the overall complexity of the model, with only a minimal reduction in accuracy compared to the baseline model. This complexity reduction translates into reduced
latency when deploying the pruned models on an NVIDIA JETSON TX2 embedded AI device.
% The abstract should briefly summarize the contents of the paper in
% 150--250 words.

\end{abstract}

\section{Introduction}

\subsection{Overview}

In the last years, Convolutional Neural Networks (CNNs) have achieved significant performance in various tasks, including Object Detection \cite{10.5555/3157096.3157139}, Image Classification \cite{NIPS2012_c399862d, simonyan2015deep, He_2016_CVPR}, Semantic Segmentation \cite{long2015fully}, and Image Captioning \cite{xu2016show}. These advancements require models to become more complex to capture intricate patterns in the data. However, large and complex models consume substantial memory and computation time. For example, VGG16 \cite{simonyan2015deep}, with its 16 layers and over 130 million parameters, requires 96MB of memory storage per image for a single forward pass, and this requirement doubles when considering the backward pass. Another example is ResNet50 \cite{He_2016_CVPR}, which requires 29 hours of training on ImageNet \cite{russakovsky2015imagenet} using an Nvidia Tesla P100 \cite{goyal2018accurate}. \\

Large and complex models are typically used to predict an output using new data, a computational process known as DNN \textit{inference}. Due to their considerable size and complexity, inference is usually performed in the cloud. In this scenario, data generated by the edge device is sent to the cloud for inference, and the result is sent back to the edge device. This process of transferring data back and forth increases latency, which can be problematic for real-time applications. To address this issue, research indicates that deploying models on edge devices for inference can reduce latency and result in faster inference times \cite{lin2021demand}. \\

Deploying large models on edge devices is not feasible due to their limited memory and computational capacity. \textit{Model Compression} emerges as a field aimed at transforming large models into smaller ones that can achieve performance similar to the larger models. A common technique for model compression, and the one discussed in this paper, is \textbf{model pruning}. Model pruning aims to reduce the size and complexity of the model by removing redundant structures such as neurons, channels, or layers. By removing these structures, the number of parameters is reduced, resulting in a decrease in model complexity and size. In practice, channel and layer pruning stands out as the most prominent pruning techniques, and they will be the focus of this paper.
% Please note that the first paragraph of a section or subsection is
% not indented. The first paragraph that follows a table, figure,
% equation etc. does not need an indent, either.

% Subsequent paragraphs, however, are indented.

\subsection{Main Contributions}
This paper proposes a novel pruning framework that combines channel and layer pruning to achieve maximal compression while maintaining the model's predictive capacity. This will be achieved by using existing channel pruning algorithms in addition to an effective layer pruning algorithm that will be proposed in the following sections. The novel approach is not the layer or channel pruning algorithm per se but rather the concept of applying them sequentially to reduce the network's width and depth. Furthermore, the paper tries to argue that utilizing existing criteria in research and applying them in a certain way can yield better results especially in terms on reducing complexity. The framework operates in two phases: it begins by executing the channel pruning algorithm, which iteratively removes redundant channels. Next, the pruned model is passed through the layer pruning algorithm, which prunes entire blocks of layers in a one-shot manner. This pruning framework reduces both the width (channels) and depth (layers) of the network. \\

This framework is inspired by EfficientNet \cite{tan2019efficientnet}, a CNN architecture developed by Google for image classification. The key innovation of EfficientNet is its method of scaling up the network's depth and width. The depth of the network refers to the number of layers, while the width refers to the number of channels or filters in each convolutional layer. While EfficientNet focuses on \textbf{scaling up} the network, the concept of this proposed pruning framework revolves around \textbf{scaling down} the network both depthwise and widthwise through channel and layer pruning. The rationale behind this approach is that DNNs, and specifically CNNs, contain a lot of redundancies across channels and layers. For instance, ResNet-1000 achieves similar accuracy to ResNet-101, despite having significantly more layers \cite{tan2019efficientnet}. This suggests that some layers (depthwise) and channels (widthwise) may not significantly contribute to the network's output, and removing them can result in a smaller network with comparable performance. \\

In summary, this paper offers the following contributions:

\begin{itemize}
\item Proposing an effective layer pruning algorithm that prunes the network in a one-shot manner based on a generated importance score for each block of layers.
\item Combining existing channel pruning algorithm with the proposed layer pruning algorithm to create an aggressive hybrid pruning framework.
\item Empirically examining the reduction in complexity of the pruned model and latency reduction on edge devices using the proposed pruning framework.
\end{itemize}

\section{Related Work}
In this section, we will discuss two pruning techniques in research which are channel and layer pruning.

\subsection{Channel Pruning}
Channel pruning is considered one technique of structured pruning that focuses on removing structures from the network, such as channels in this case, thereby maintaining a more regular network architecture. In contrast, unstructured pruning (weight pruning) is a fine-grained technique used to remove irrelevant connections or weights from a model. Weight pruning is not very common because, even though it can significantly reduce the number of parameters, this reduction does not necessarily translate into a reduction in latency due to the irregular structure resulting from this pruning. \\

There is an extensive amount of research done on channel pruning ranging from simple to sophisticated algorithms. In Li et al.'s approach~\cite{li2017pruning}, a global pruning method is employed where all filters are ranked based on their absolute values. Filters with lower magnitudes are then removed. ThiNet, as described by Luo et al.~\cite{luo2017thinet}, adopts a data-driven strategy to select filters for pruning. It assesses the importance of each filter and considers its impact on the subsequent layer's output, ensuring effective pruning decisions. Molchanov et al.~\cite{molchanov2019importance} propose a method utilizing Taylor expansion of the first and second orders to capture the sensitivity of the loss function to any changes in the parameters. HRank, introduced by Lin et al.~\cite{lin2020hrank}, calculates the rank of filters within the network and subsequently prunes those with lower rank values, indicating their perceived insignificance to model performance.
\subsection{Layer Pruning}
Layer pruning is a structured pruning technique where unnecessary layers are systematically removed from the network. This approach, however, is less explored compared to channel pruning due to its potential aggressiveness. AutoLR, detailed in Ro et al.'s work \cite{ro2021autolr}, introduces a layer pruning algorithm that evaluates each layer's contribution to the target task. Initially, the network's performance is measured before any pruning occurs. Subsequently, layers are pruned iteratively, starting from the highest-level layers closer to the output. If the pruned model exhibits improved performance compared to its initial state, the pruning process continues, updating the performance benchmark accordingly. Conversely, if performance declines, the pruning decision is reconsidered, allowing the network to revert to its initial configuration. DepthShrinker, proposed by Facebook in Fu et al.~\cite{pmlr-v162-fu22c}, targets the reduction of redundant activations within network blocks. The algorithm achieves this by removing unnecessary activations and subsequently merging consecutive layers into a single compact and dense layer.

\section{Methodology}
This section will explain the underlying techniques used in the proposed pruning framework. Our hybrid pruning framework consists of two phases: First, existing channel pruning algorithms are used to reduce the width of the network. Second, our proposed layer pruning algorithm is used to reduce the depth of the network. Both channel and layer pruning will be using the same criteria for pruning. Section~\ref{Pruning-Criteria} will discuss the various criteria used in our framework and Section~\ref{Layer-Pruning-Algorithm} will explain the proposed layer pruning algorithm.

\subsection{Pruning Criteria}
\label{Pruning-Criteria}
Research in model pruning has devised many criteria for pruning ranging from simple to sophisticated ones. The main objective is to develop criteria that is able to identify unimportant components in the model for pruning without jeopardizing the performance. However, there are no optimal criteria that work all the time, and existing criteria are heuristic. Therefore, the objective of this section and the upcoming ones is to experiment with various criteria in our framework and decide on the best one empirically. The following criteria are primarily proposed for channel pruning. However, we generalize them to apply to our proposed layer pruning. This generalization is achieved by averaging the scores of all the filters in a layer to obtain a score for that specific layer. Hence, the following criteria will be utilized in both the channel and layer pruning algorithms inside our hybrid pruning framework.

\subsubsection{Weight Magnitude:}
In CNNs, weight magnitude~\cite{li2017pruning} is a simple yet effective method for identifying unimportant components. It calculates the absolute value of the weights and considers those with higher values to be more important than other weights. A low weight magnitude indicates that the produced feature map contains weak activations, meaning the filter does not contribute significantly to the overall output. Hence, components with low weight magnitude are pruned. This criterion evaluates the weights in a specific filter and to generalize it to the whole layer, we take the average of all the filters in a specific layer. The weight magnitude importance per layer can be computed as follows: 
\begin{equation}
        \text{WM\_I = }
        \dfrac{1}{c^{[l]}} \sum_{i=0}^{c^{[l]}} \norm{W^{[l]}[:,i,:,:]}
\label{eqn:weightMagnitude}
\end{equation}
$W^{[l]}$ is the weight matrix at layer $l$. Matrix $W^{[l]} \in R^{N^{[l]} * c^{[l]} * f^{[l]} * f^{[l]}}$ where $c^{[l]}$ is the number of filters at layer $l$, $N^{[l]}$ is the input channels at layer $l$, and $f^{[l]}$ is the filter size of the channel.

\subsubsection{Batch Normalization Scale:}
The batch normalization layer is a layer used in neural networks and CNNs to improve the overall training process. This method introduces a scaling factor $\gamma$ in the batch normalization layer and applies L1 regularization on the scaling factors during training. This regularization encourages these scaling factors to become or approach zero which indicates that the corresponding channel may be pruned~\cite{liu2017learning}. This results in the pruning of significant channels resulting in a thinner and compact model. The batch normalization scale can be used to estimate the importance of layers using the following formula: 

\begin{equation}
        \text{BN\_I = }
        \dfrac{1}{c^{[l]}} \sum_{i=0}^{c^{[l]}} (\gamma^{[l]})^2
\label{eqn:BNscale}
\end{equation}
where $\gamma^{[l]}$ is the Batch Normalization Scale Factor at layer $l$

\begin{figure}
\includegraphics[width=\textwidth]{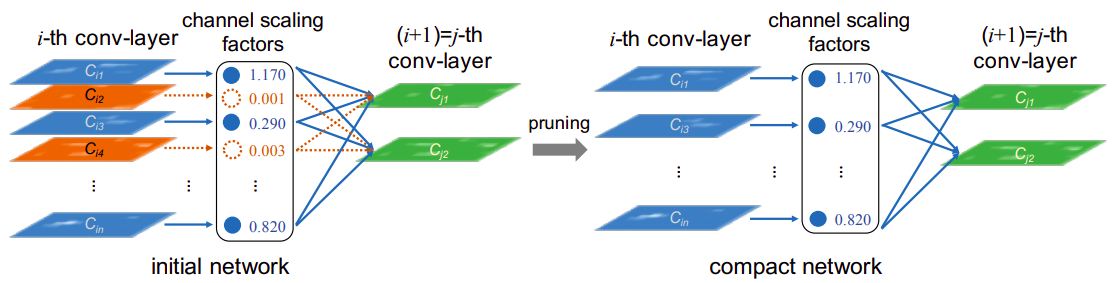}
\caption{Visual illustration of Batch Normalization Scale. Each channel has
a scaling factor (from the batch normalization layer). The channel with a small scaling factor (orange color) will be pruned~\cite{liu2017learning}} \label{fig1}
\end{figure}

\subsubsection{Feature Map Rank:}
This method introduces the use of the rank of the feature maps which is the number of non-zero singular values as a criterion for channel pruning~\cite{lin2020hrank}. The algorithm calculates the rank of feature maps, ranks the channels accordingly, and prunes those with the lowest rank. The importance score using feature map rank can be calculated as follows:

\begin{equation}
        \text{FMR\_I = }
        \dfrac{1}{c^{[l]}} \sum_{i=0}^{c^{[l]}} rank(F_{i})
\label{eqn:featureMap}
\end{equation}
where $F_{i}$ is the feature map produced by the $i$-th filter and $rank(F_{i})$ is the rank of the feature map produced by the $i$-th filter. The rank of the feature map $F_{i}$ is computed as follows:

\begin{equation}
        rank(F_{i}) = \sum_{j}1(\sigma_j > \epsilon)
\label{eqn:rank}
\end{equation}
where $\sigma_j$ is the singular values obtained from the SVD of $F_{i}$ and $1(.)$ is an indicator function that counts values greater than the threshold $\epsilon$. In our case, a value of $10^{-3}$ has been used for $\epsilon$.

\subsubsection{Taylor Method:}
This method estimates the importance of each filter in CNN using the first-order Taylor expansion of the loss function~\cite{molchanov2019importance}. The main idea is to estimate the importance of each parameter by simply removing them and calculating the error it incurs upon removal. However, computing the induced error set for each parameter is computationally demanding, as it necessitates evaluating multiple network versions, one for each removed parameter. The idea of the Taylor method is to estimate the induced error set by approximating it around W using the first-order Taylor expansion which can approximate the change in the loss function when a filter is removed. It can be computed as follows:

\begin{equation}
        \text{TM\_I = }
        \dfrac{1}{c^{[l]}} \sum_{i=0}^{c^{[l]}} \bigg | \frac{\partial L}{\partial W_{i}} W_{i} \bigg |
\label{eqn:Taylor}
\end{equation}
where $L$ is the loss function and $\frac{\partial L}{\partial W_{i}}$ is the gradient of the loss w.r.t the weights of the $i$-th filter

\subsection{Layer Pruning Algorithm}
\label{Layer-Pruning-Algorithm}
The motivation behind developing a new algorithm for layer pruning, rather than focusing on channel pruning, stems from the abundance of research on channel pruning, which encompasses a wide spectrum of algorithms. In contrast, layer pruning has not received much attention, with few effective algorithms developed for this purpose. Therefore, one of the objectives of this paper is to devise a layer pruning algorithm that utilizes established criteria and prunes layers in a one-shot strategy. The algorithm draws significant inspiration from the work of Molchanov et al.~\cite{molchanov2019importance}. \\

\begin{figure}[t]
\begin{center}
\includegraphics[width=0.8\textwidth]{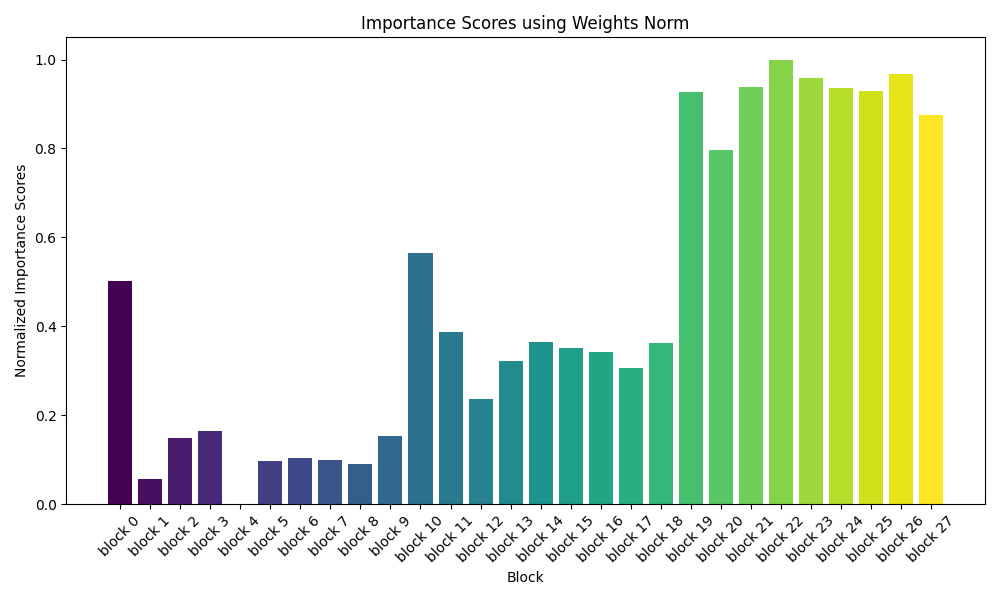}
\caption{Histogram showing the importance score for ResNet56 using Weight Magnitude. It should be noted that ResNet56 has 3 groups and each consists of 9 residual blocks. These residual blocks contain 2 convolutional layers.} \label{fig2}
\end{center}
\end{figure} 
The algorithm starts by calculating the importance score for each block of layer using the criteria mentioned above. All the scores are generated before the pruning starts. Figure~\ref{fig2} shows the generated importance score using the weight magnitude criterion. The next step is to sort the blocks based on their importance score. After sorting, pruning starts in a one-shot manner which means that the N (where N is given as an argument) least important blocks are pruned in a single pass. Finally, the pruned model is fine-tuned for weight and performance adjustment. \\

This algorithm is effective as it leverages concepts such as importance estimation and one-shot pruning. Additionally, we have experimented with several criteria to determine the optimal ones and gain a deeper understanding of their functionality within our pruning framework. At this stage, the two main components of the pruning framework are ready and it is time to combine channel and layer pruning. There are two implementations of the framework, starting with channel pruning and then following it with layer pruning and vice versa. Empirically, we conclude that starting with channel pruning and then layer pruning yields better results than starting with layer pruning and following it with channel pruning. Hence, in the experiments section, we will only present the results of our pruning framework where we start with channel pruning followed by layer pruning. The results of the pruning framework when starting with layer pruning followed by channel pruning will be found in Table~\ref{table3} and~\ref{table4} in the Appendix section.

\begin{figure}[t]
\begin{center}
\includegraphics[width=0.5\textwidth]{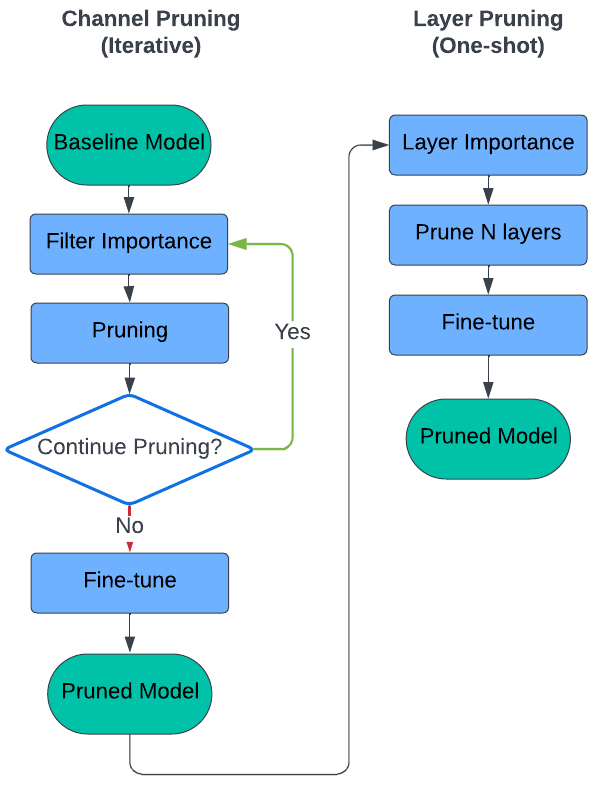}
\caption{Visual illustration of the proposed pruning framework} \label{fig3}
\end{center}
\end{figure}

\section{Experiments}
In this section, the results obtained from the pruned model will be presented. First, the implementation details used in the experiments for both pruning and inference will be discussed. Then, the results of the pruning framework approach will be presented and briefly analyzed.
\subsection{Experimental Setup}
The experiments were carried out on IDUN \cite{sjalander+:2019epic}, a GPU cluster coordinated by NTNU’s faculties and the IT division, providing a high-availability computing platform. Specifically, using NVIDIA P100, this environment is utilized for pruning ResNet-56 multiple times across various criteria using the hybrid approach. The hybrid approach, involving training, pruning, retraining, and evaluating performance took approximately 4 hours to complete for 5 criteria. This process has been iterated multiple times, exploring variations such as layer pruning followed by channel pruning, and vice versa. Additionally, the process was repeated for various pruning ratios. Consequently, most of the time was dedicated to experimenting with different methods to gather comprehensive results. \\

The edge device chosen for this paper is the NVIDIA Jetson TX2, which is commonly used and ideal for intelligent edge applications such as robots, drones, and smart cameras. It features a 256-core NVIDIA Pascal GPU with 256 NVIDIA CUDA cores. These GPU cores were utilized during the process of running the pruned models on the Jetson TX2. In future work, the NVIDIA Jetson Nano will be used as it offers more features and is much easier to set up compared to the TX2. The main objective is to evaluate the performance of the pruned models on any embedded devices so the choice between the various NVIDIA boards is not a priority now. After obtaining the pruned models, the next step is to run them on the edge device to measure latency. Latency is measured by manually timing the inference time using high-resolution timers. Latency results are then averaged over 100 forward passes, preceded by a few warm-up forward passes for lazy GPU initialization. It should be noted that all work from pruning to evaluation has been implemented using PyTorch~\cite{paszke2019pytorchimperativestylehighperformance}. It should be also noted that several optimizations could enhance our results, such as replacing PyTorch implementations with TensorRT, which is available on NVIDIA hardware. However, the primary aim of this paper is to present a new pruning framework and evaluate the results of our model within this framework. In the future, we can consider optimizing various aspects of our framework to improve results and maximize performance by leveraging specific hardware features on NVIDIA GPUs. The interaction with the Jetson device was conducted through a headless setup (without a monitor) on Ubuntu 18.04.

\subsection{Experimental Results}
This section presents the results obtained from our pruning framework. First, channels are pruned to reduce the network’s width by removing redundant channels or filters. Next, the layer pruning algorithm explained in Section~\ref{Layer-Pruning-Algorithm} prunes redundant layers or blocks, thereby reducing the network’s depth. Figure~\ref{fig3} provides a visual illustration of the overall framework. The results shown below are obtained using ResNet-56 with the default configuration and hyperparameters on CIFAR-100~\cite{Krizhevsky2009LearningML}. These results document three metrics: accuracy, complexity, and latency based on the previously discussed criteria. The complexity of the model is expressed in the number of parameters and floating point operations (FLOPs)\\

Tables~\ref{table1} and~\ref{table2} summarize the accuracy, complexity, and latency of the models after pruning using our framework. The main objective is to strike a balance between these three metrics, aiming to achieve a significant reduction in both complexity and latency while maintaining a satisfactory accuracy compared to the baseline model. \\

\newpage
\setcounter{footnote}{0}

\begin{table}[t]
\centering
\caption{The table summarizes the accuracy and the complexity of each pruned model according to each criterion. Best results are shown in \textbf{bold}}
\label{table1}
\begin{tabular}{l|c|c|c}
\toprule
Method &  Accuracy  &   Parameters  &   FLOPs \\
\midrule
Baseline ResNet-56 on CIFAR-100  &  71.28 \%    &   858,868    &  127,621,440\\
Weight Magnitude (1 block)\tablefootnote{This means 1 block is pruned using Layer pruning algorithm}          &  69.48 \%   & 803,900   & 93,747,360 \\
Weight Magnitude (2 blocks)\tablefootnote{This means 2 blocks are pruned using Layer pruning algorithm}         &  69.19 \%     & 801,866   & 91,750,560\\
Batch Normalization Scale (1 block)  &  69.87 \%    &   743,578     &   76,024,992 \\
Batch Normalization Scale (2 blocks) &  69.68 \%    &   741,348     &   \textbf{73,797,792}\\
Feature Maps Rank (1 block)          &  \textbf{70.67} \%    &   604,326     &   91,881,112 \\
Feature Maps Rank (2 blocks)         &  70.11 \%    &   \textbf{559,850}     &   89,005,336\\
Weight Taylor (1 block)              &  69.64 \%    &   799,810     &   93,563,040 \\
Weight Taylor (2 blocks)             &  69.60 \%    &   797,120     &   90,902,688\\
\bottomrule
\end{tabular}
\end{table}

Table~\ref{table1} presents the accuracy and complexity of the pruned models compared to the baseline ResNet-56. It is expected to have a reduction in accuracy since the models undergo a two-phase pruning process. From the table, we can observe that the Feature Map Rank (1 block) achieves the best accuracy results, with less than a 1\% reduction. Since a block contains two convolutional layers, the Feature Map Rank model, with two layers and many channels pruned, manages to achieve remarkable results. Additionally, there is a noticeable reduction in the number of parameters and FLOPs in almost all the models, which is intuitive as they undergo an aggressive pruning technique. This reduction in parameters will lead to a decrease in memory consumption, which is critical for edge devices. Similar to accuracy, the Feature Map Rank with 2 blocks achieves the best reduction in the number of parameters, with approximately 300,000 parameters removed compared to the baseline model. Regarding FLOPs, the batch normalization scale with 2 blocks achieves the best reduction, with more than 50 million operations removed. It is clear from the table that the models undergo a significant reduction in complexity while maintaining a relatively stable accuracy.\\

The last metric to consider is latency, and Table~\ref{table2} shows the latency and latency reduction of the pruned models. Unlike the previous two metrics, neither Feature Map Rank nor Batch Normalization Scale achieves the best latency reduction. Weight Magnitude with 1 block removed achieves the best latency reduction. This is a very interesting observation that will be further explained in the discussion section, but it seems that a significant reduction in complexity might not always translate into a reduction in latency. Weight Magnitude did not achieve the best results in accuracy or complexity, yet it stands as the best in latency reduction.\\

In the appendix, Tables \ref{table3} and \ref{table4} present the results for the second implementation of the hybrid pruning framework, which begins with pruning layers followed by channels. This implementation performs worse than the previously discussed one in terms of accuracy, complexity, and latency. We did not delve into the details of this implementation and focused instead on the first one, which starts with channel followed by layer pruning, as it yields better results. \\

\begin{table}[t]
\centering
\caption{The table summarizes the latency and latency reduction of each pruned model according to each criterion. Best results are shown in \textbf{bold}}
\label{table2}
\begin{tabular}{l|c|c}
\toprule
Method &  Latency   & Latency Reduction \\
\midrule
Baseline ResNet-56 on CIFAR-100      &  61.207 ms   &   -             \\
Weight Magnitude (1 block)           &  \textbf{34.338 ms}   &   \textbf{43.89} \%      \\
Weight Magnitude (2 blocks)          &  34.852 ms   &   43.05 \%      \\
Batch Normalization Scale (1 block)  &  35.051 ms   &   42.73 \%      \\
Batch Normalization Scale (2 blocks) &  34.783 ms   &   43.17 \%      \\
Feature Maps Rank (1 block)          &  39.373 ms   &   35.67 \%      \\
Feature Maps Rank (2 blocks)         &  37.902 ms   &   38.07 \%      \\
Weight Taylor (1 block)              &  34.652 ms   &   43.38 \%      \\
Weight Taylor (2 blocks)             &  34.865 ms   &   43.03 \%      \\
\bottomrule
\end{tabular}
\end{table}

\section{Discussion}
The results showed various patterns among the three metrics. However, there is a noticeable reduction in both complexity and latency, while accuracy has only slightly degraded. This demonstrates that our pruning framework operates aggressively, which can be useful in scenarios where memory is extremely limited. On the other hand, there are some discrepancies in the results, where some models achieve good latency reduction but not a significant reduction in complexity. \\

\begin{figure}[t]
\begin{center}
\includegraphics[width=0.8\textwidth]{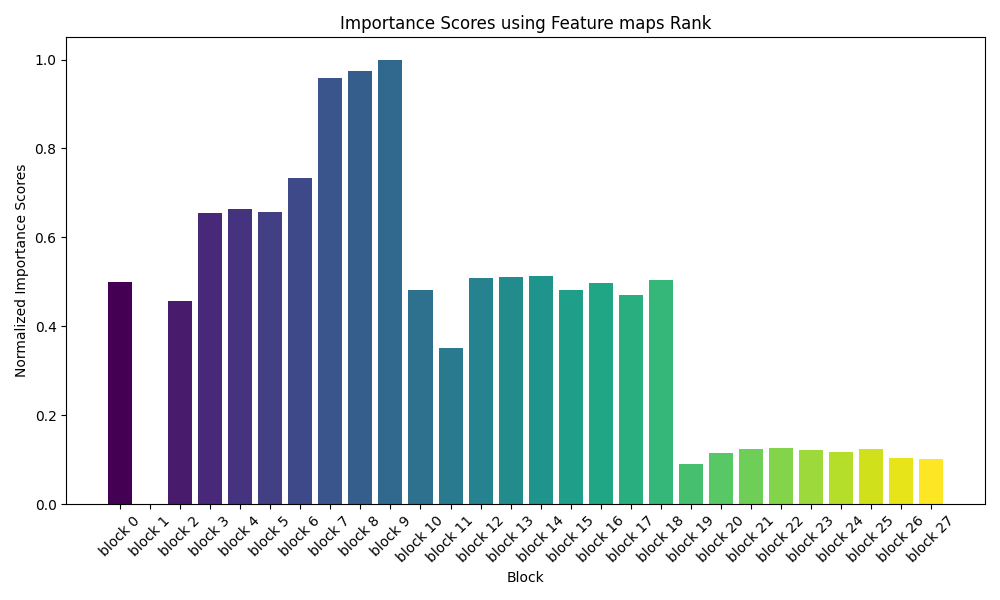}
\caption{Histogram showing the importance score for ResNet56 using Feature Map Rank} \label{fig4}
\end{center}
\end{figure}

To better understand how pruning works, we will focus on two criteria from Table~\ref{table1}: \textit{Weight Magnitude} and \textit{Feature Map Rank}. In Figure~\ref{fig2}, we see that Weight Magnitude assigns high importance to deeper layers while assigning less importance to earlier layers. Conversely, Figure~\ref{fig4} shows that Feature Map Rank assigns more importance to the earlier layers while assigning less importance to the deeper ones. Feature maps are known to become sparser as we go deeper into the network, hence, Feature Map Rank assigns a lower importance score to deeper layers. Consequently, Feature Map Rank tends to prune deeper layers, considering them less important, and deeper layers usually have more parameters due to the increased number of channels. On the other hand, Weight Magnitude works in the opposite direction by tending to prune earlier layers, as it considers them less important. This explains why, in Table~\ref{table1}, Feature Map Rank achieves a better reduction in complexity compared to Weight Magnitude. \\

The same rationale applies to the accuracy recorded in Table~\ref{table1}. \textit{Feature Map Rank} primarily prunes deeper layers, which typically has minimal impact on overall accuracy since these layers capture highly specific patterns. Removing them may not disrupt the flow of information in the network. Conversely, \textit{Weight Magnitude} tends to prune earlier layers, potentially affecting performance because these layers and channels are vital for processing input data. The evaluation of which criterion is superior depends on the specific use case.\\

The latency results shown in Table~\ref{table2} reveal an intriguing observation. Despite \textit{Feature Map Rank} achieving remarkable outcomes in both accuracy and complexity, it does not demonstrate substantial latency reduction compared to other criteria. One plausible explanation for this phenomenon is that aggressive pruning of deeper layers might lead to memory access issues. This can result in irregular memory accesses, causing numerous cache misses and inefficient memory bandwidth utilization, thereby hindering optimal latency reduction for the model. Table~\ref{table2} illustrates that \textit{Weight Magnitude}, which underwent less aggressive pruning compared to other criteria, achieves the most significant latency reduction. This suggests that there may exist a pruning threshold beyond which latency performance begins to degrade. Further investigation into the pruned network's topology and structure is necessary to grasp the trade-offs between complexity and latency reduction. \\

The primary objective of this paper is to achieve a harmonious balance between accuracy, complexity, and latency. Although seemingly straightforward, these metrics are intricately interlinked, where even a slight alteration in one can be reflected across the others. It is well-established that a more complex model tends to yield better accuracy. However, a complex model results in an increase in latency and this is not desirable when implemented on embedded devices as these devices need lower latency for real-time processing. Hence, balancing these metrics looks like a vicious circle. This paper aims at reducing the model’s complexity, thereby reducing latency, while simultaneously trying to avoid a degradation in the accuracy. This is done by starting the pruning process in an iterative process and fine-tuning the model carefully. Additionally, careful selection of redundant components using sophisticated pruning criteria can help maintain accuracy. In conclusion, there are several trade-offs between the metrics, and striking a balance between them is the goal when it comes to model compression. 

% This paper aims to contribute to the research community by introducing a novel model compression technique that integrates both channel and layer pruning. While each of these methods has been extensively explored in literature, there has been little discussion on combining them to achieve maximal compression, which can prove beneficial in certain scenarios. While this paper is not without its imperfections, it offers a fresh perspective in the field of model compression research. \\

\section{Conclusion}
This paper aims to contribute to the research community by introducing a novel model compression framework that integrates both channel and layer pruning. The aim of this hybrid pruning framework is to achieve maximal compression for CNNs without undermining the model's predictive capacity. The framework evaluates its performance based on three metrics: accuracy, complexity, and latency. \\

Based on the results in Table~\ref{table1} and~\ref{table2}, it is feasible to implement an aggressive pruning technique while retaining the predictive capacity of the model. This is achieved by adapting EfficientNet’s idea of scaling up the network depthwise and widthwise to make the model bigger. In this paper, the network is scaled down depthwise (layer pruning) and widthwise (channel pruning), resulting in a significant reduction in both the number of parameters and FLOPs. Although the accuracy of the pruned models showed some variation, they generally performed well considering the aggressive pruning applied. Additionally, there was a noticeable reduction in latency across nearly all pruned models. This demonstrates that applying different pruning methods, such as channel and layer pruning, can achieve significant reductions in model size and complexity. \\

This aggressive hybrid pruning framework is especially crucial when applied to real-world applications, particularly those running on edge devices such as smartphones, IoT devices, or embedded systems. These devices often face limitations in computational power, memory, and storage. Furthermore, real-time applications like autonomous driving demand rapid response times, which can be achieved by reducing model latency. Our framework addresses these challenges by streamlining models to run efficiently on constrained hardware, eliminating bottlenecks that traditionally undermine AI deployment in practical scenarios. By reducing the model's size, complexity, and latency, our framework makes AI technologies more accessible, efficient, and viable for widespread use across various real-world applications. \\

In conclusion, the main findings of our paper can be summarized as follows:
\begin{itemize}
\item The paper proposes a novel pruning framework that integrates channel and layer pruning to compress the network both depthwise and widthwise.
\item It introduces an effective layer pruning algorithm that builds upon existing channel pruning criteria.
\item The paper experiments with existing channel pruning algorithms and incorporates them into its framework.
\item Several pruning criteria are explored in the research to test the capabilities of the proposed framework.
\item The performance of the pruning framework is evaluated using three distinct metrics: accuracy, complexity, and latency.
\end{itemize}

\begin{credits}
\subsubsection{\discintname}
The authors have no competing interests to declare that are relevant to the content of this article
\end{credits}

\bibliographystyle{splncs04}
\bibliography{mybibliography}

\newpage
\section{Appendix}
\subsection*{\large{Additional Tables}}

\begin{table}
\centering
\caption{The table summarizes the accuracy and the complexity of each pruned model according to each criterion. The results were obtained after applying layer pruning followed by channel pruning. Best results are shown in \textbf{bold}}
\label{table3}
\begin{tabular}{l|c|c|c}
\toprule
Method &  Accuracy  &   Parameters  &   FLOPs \\
\midrule
Baseline ResNet-56 on CIFAR-100  & 71.28 \% & 858,868  & 127,621,440\\
Weight Magnitude (1 block)       & 68.48 \% & 833,326  & 109,383,824\\
Weight Magnitude (2 blocks)      & 68.13 \% & 824,778  & 105,401,488\\
Batch Normalization Scale (1 block)& \textbf{68.67} \% &740,236 & 78,915,736 \\
Batch Normalization Scale (2 blocks) &  68.08 \%    &   733,917     &   \textbf{78,666,896}\\
Feature Maps Rank (1 block)          &  65.70 \%    &   591,486     &   91,576,216 \\
Feature Maps Rank (2 blocks)         &  65.36 \%    &   \textbf{587,394}     &   91,560,088\\
Weight Taylor (1 block)              &  68.64 \%    &   822,234     &   106,508,440 \\
Weight Taylor (2 blocks)             &  67.87 \%    &   819,460     &   104,858,384 \\
\bottomrule
\end{tabular}
\end{table}

\begin{table}
\centering
\caption{The table summarizes the latency and latency reduction of each pruned model according to each criterion. The results were obtained after applying layer pruning followed by channel pruning. Best results are shown in \textbf{bold}}
\label{table4}
\begin{tabular}{l|c|c}
\toprule
Method &  Latency   & Latency Reduction \\
\midrule
Baseline ResNet-56 on CIFAR-100  &  61.207 ms   &   -       \\
Weight Magnitude (1 block)       &  35.723 ms   &  41.63 \%      \\
Weight Magnitude (2 blocks)          &  35.977 ms   &   41.22 \%      \\
Batch Normalization Scale (1 block)  &  35.880 ms   &   41.38 \%      \\
Batch Normalization Scale (2 blocks) &  \textbf{34.344 ms}  &   \textbf{43.89} \%      \\
Feature Maps Rank (1 block)          &  34.721 ms   &   43.29 \%      \\
Feature Maps Rank (2 blocks)         &  35.991 ms   &   41.20 \%      \\
Weight Taylor (1 block)              &  38.032 ms   &   37.86 \%      \\
Weight Taylor (2 blocks)             &  37.075 ms   &  39.43 \%      \\
\bottomrule
\end{tabular}
\end{table}

\end{document}